\documentclass[journal]{IEEEtai}
\usepackage[colorlinks,urlcolor=blue,linkcolor=blue,citecolor=blue]{hyperref}
\usepackage{color,array}
\usepackage{graphicx}
\usepackage{cite}
\usepackage{amssymb}
\usepackage[T1]{fontenc}
\usepackage[utf8]{inputenc}
\usepackage{authblk}
\usepackage{float}
\usepackage{booktabs}
\usepackage{subfigure}
\usepackage{amsmath}
\usepackage{multirow}
\usepackage[normalem]{ulem}
\usepackage{caption}
\useunder{\uline}{\ul}{}
\usepackage[linesnumbered, ruled]{algorithm2e}
\SetKwRepeat{Do}{do}{while}%
\usepackage{abstract}

\begin{document}

\title{HICH Image/Text (HICH-IT): Comprehensive Text and Image Datasets for Hypertensive Intracerebral Hemorrhage Research}

\author[2]{Jie Li}
\author[1]{Yulong Xia}
\author[2]{Tongxin Yang}
\author[2]{Fenglin Cai}
\author[3]{Miao Wei}
\author[3]{Zhiwei Zhang}
\author[1]{Li Jiang*\thanks{*Corresponding author:Li Jiang,  drjiangli2019@163.com}}

\affil[1]{Department of Neurosurgery, the First Affiliated Hospital of Chongqing Medical University, Chongqing, PR China}
\affil[2]{Chongqing University of Science and Technology, Chongqing, PR China}
\affil[3]{Department of Radiology, the First Affiliated Hospital of Chongqing Medical University, Chongqing, PR China}

\maketitle

\begin{abstract}
In this paper, we introduce a new dataset in the medical field of hypertensive intracerebral hemorrhage (HICH), called HICH-IT, which includes both electronic medical records (EMRs) and head CT images. This dataset is designed to enhance the accuracy of artificial intelligence in the diagnosis and treatment of HICH. This dataset, built upon the foundation of standard text and image data, incorporates specific annotations within the EMRs, extracting key content from the text information, and categorizes the annotation content of imaging data into four types: brain midline, hematoma, left and right cerebral ventricle. HICH-IT aims to be a foundational dataset for feature learning in image segmentation tasks and named entity recognition. To further understand the dataset, we have trained deep learning algorithms to observe the performance. The pretrained models have been released at both www.daip.club and github.com/Deep-AI-Application-DAIP. The dataset has been uploaded to https://github.com/CYBUS123456/HICH-IT-Datasets.

Index Terms-HICH, Deep learning, Intraparenchymal hemorrhage, named entity recognition, novel dataset
\end{abstract}

\section{Introduction}
Although spontaneous intracerebral hemorrhage (ICH) accounts for less than 20\% of all stroke cases, it remains the subtype with the highest mortality and morbidity rates among all forms of cerebrovascular accidents\cite{gross2019cerebral,hostettler2019intracerebral,qureshi2001spontaneous,fewel2003spontaneous}. Hypertensive intracerebral hemorrhage (HICH) as one of the most common ICH poses a significant threat to patients, and with limited treatment options available, the condition not only becomes a burden for the families of those affected but also presents challenges to national healthcare systems. HICH develops rapidly and has a poor prognosis. It requires prompt and accurate diagnosis and treatment to avoid irreversible consequences for the patients. Therefore, the rapid and accurate identification of key clinical manifestation and analysis of head CT scans are crucial to the diagnosis and treatment \cite{chen2023review,thrift1996risk,fisher1971pathological,brott1986hypertension,magid2022cerebral}.

Despite the advanced nature of current methodologies in head CT and text extraction, challenges remain in terms of precision and contextual understanding\cite{yi2019generative,lundervold2019overview,zhou2021review,campos2020yake,wang2018clinical}. Meanwhile, there is a scarcity of meticulously annotated head CT and text datasets related to HICH, resulting in a lack of reliable data support and reserve for the field.

HICH-IT is the first Chinese database for HICH. And is dedicated to support open research in the fields of head CT image segmentation and EMR extraction for patients with HICH:
\begin{enumerate}[]
\item We provide a dataset of HICH images and EMR named entities.
\item The dataset is comprehensively described in terms of its origin, composition, format, and labels.
\item The performance of the dataset on deep learning models is presented.
\end{enumerate}

\begin{figure*}
	\centering
	\includegraphics[width=16cm,height=9cm]{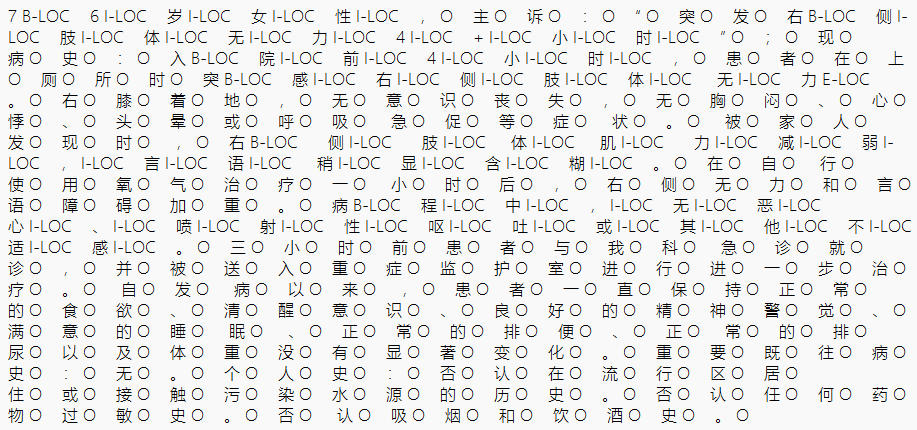}
	\caption{Partial of the text data are provided, due to the extensive volume of text information, a selection of this data is displayed for illustrative.}
	\label{Fig.text}`
\end{figure*}

\section{Related Works}
The dataset originates from authentic medical records, in order to protect patient privacy, we have excluded sections that are closely related to patient privacy from the original records. The dataset consists of two distinct annotated datasets: one for head CT images and the other for EMRs. The image data, annotated using 3D-Slicer, is segmented into NIfTI format slice images. The annotation data for the images consists of four parts: the hematoma, left cerebral ventricle, right cerebral ventricle, and the Brain middle. The positional information of all four parts has been comprehensively annotated. The text data includes detailed text information related to the patient's medical history, current condition, physical examination findings, laboratory results, and examination reports, all of which are closely linked to subsequent diagnosis and treatment. These information has been manually annotated using methods based on the BERT model. This is done to facilitate text information processing approaches such as Named Entity Recognition (NER).

This dataset plays a crucial role in interdisciplinary research, particularly in the fields of medicine, computer science, and artificial intelligence. With the limited availability of extensively annotated datasets for HICH head CT images and EMRs, our objective is to create this dataset to drive progress in intelligent healthcare. The dataset enables experimentation with models for image segmentation and text recognition. Moreover, it helps clinical professionals quickly understand patient conditions, ultimately improving diagnostic and treatment effectiveness. In the field of artificial intelligence, this dataset provides a wealth of experimental data for the development of more advanced medical image processing and EMR analysis algorithms. This contributes to achieving more accurate and effective methods for medical diagnosis and treatment\cite{shen2017deep,greenspan2016guest,litjens2017survey,erickson2017machine}.

\begin{figure*}
	\centering
	\subfigure[Hematoma]{
		\label{fig.blood}
		\includegraphics[width=4cm,height=4cm]{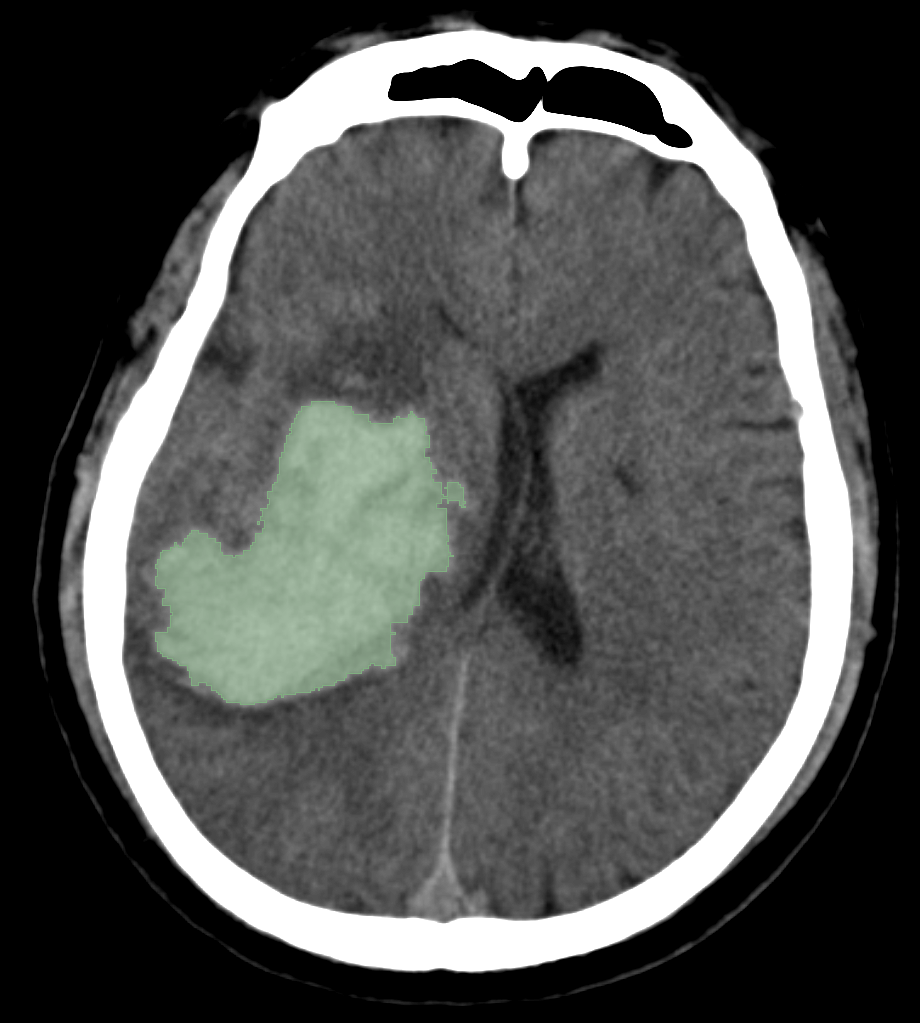}}
	\subfigure[Brain midline]{
		\label{fig.middle}
		\includegraphics[width=4cm,height=4cm]{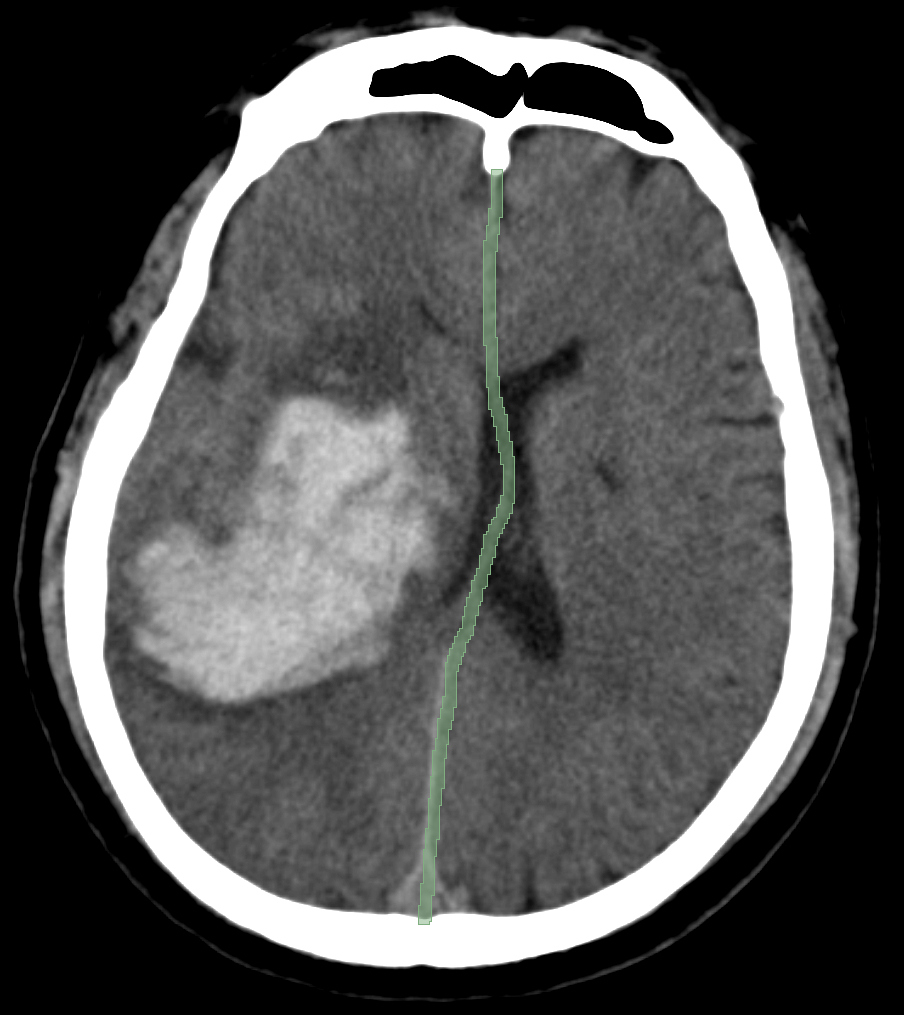}}	\subfigure[Left lateral ventricle]{
		\label{fig.left}
		\includegraphics[width=4cm,height=4cm]{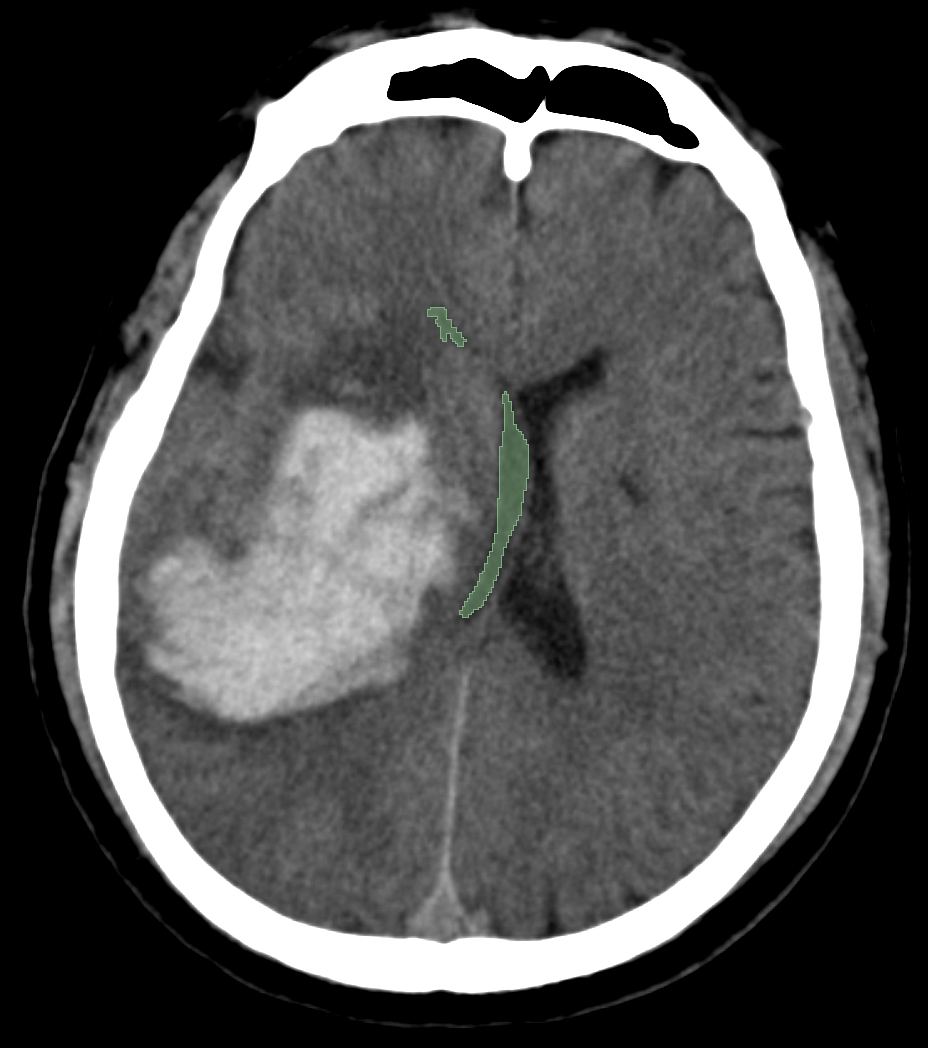}}
	\subfigure[Right lateral ventricle]{
		\label{fig.right}
		\includegraphics[width=4cm,height=4cm]{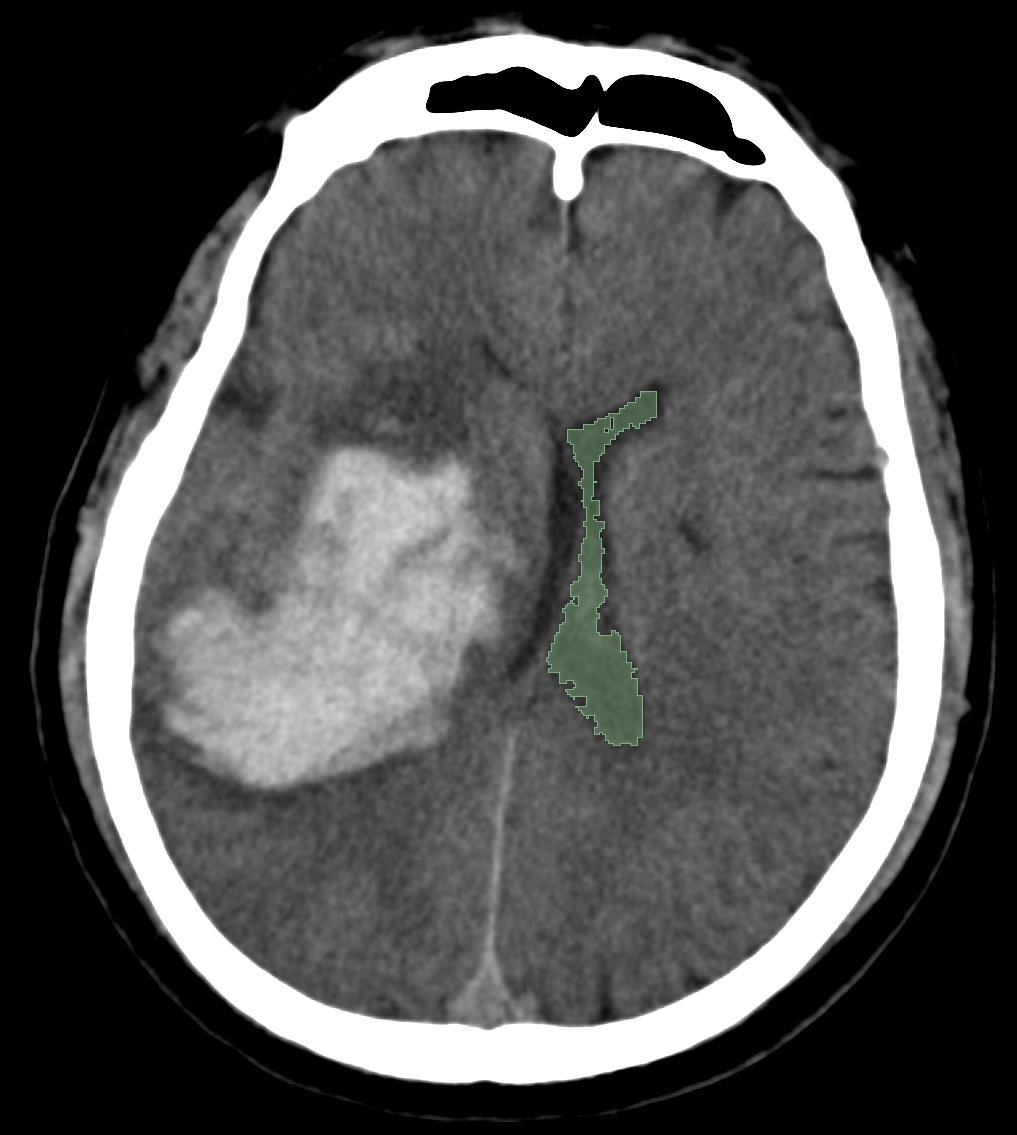}}
	\caption{Head image. (a) Displays the hematoma annotation; (b) Displays the brain middle; (c) Displays the left cerebral ventricle annotation; (d) Displays the right cerebral ventricle annotation. All these images represent the viewing effects within 3D-Slicer.}
	\label{Fig.main}
\end{figure*}

\section{HICH-IT DATASET}
\subsection{Data Source}
The foundation of any impactful medical research lies in the quality and relevance of the data sources used. Our objective is to utilize and continuously update this dataset for the long term. Therefore, the source of the data is of paramount importance. The EMR and head CT image data in the HICH-IT are sourced from the emergency electronic medical record database of several medical centers. The image data consists of head CT scans completed during patients's visits to the emergency department, while the text data consists of text data in electronic medical records.

\subsection{Dataset Composition}
This dataset encompasses several hundred thousand CT images and thousands of case texts. Additionally, due to the extensive volume of text information, we have selected some of the information to show, the images are represented as viewed within 3D-Slicer. This dataset is composed of head CT images and EMR texts,  both carefully annotated. The annotation information for the head CT images is divided into four categories: hematoma, midline, left cerebral ventricle, and right cerebral ventricle, as illustrated in Figure\ref{Fig.main}. The case text information has been annotated for significant sections using methods based on the BERT model, as depicted in Figure\ref{Fig.text}. In addition, since the text data is stored as a whole column in the text file, it is necessary to make formatting adjustments in order to achieve a better display effect.
\subsection{Data Format}
Due to that medical image data comprises four key com-ponents: pixel depth, photometric interpretation, metadata, and pixel data. Therefore, the preferred format for storing medical datasets is typically the NIfTI format. NIfTI images are often three-dimensional, representing   sagittal, coronal, and axial planes upon slicing. The advantage of this format lies in its ability to accurately reflect metadata, including directional in-formation, making it highly suitable for neurosurgical-related image data\cite{larobina2014medical,tournier2019mrtrix3,willemink2020preparing}. Our dataset's head image data consists of several hundred thousand CT images, which are stored in NIfTI format. Each image slice measures 512*512 pixels. The text data, on the other hand, comprises tens of thousands of cases and is stored in TXT format.

\begin{table*}[h]
	\begin{center}
		\begin{minipage}{\textwidth}
			\caption{Text experiment results.}\label{tab1}
			\begin{tabular*}{\textwidth}{@{\extracolsep{\fill}}lcccccc@{\extracolsep{\fill}}}
				\toprule
				Parameters & Precision & Recall & F1-score & Support \\
				\midrule
				O & 0.96 & 0.98 & 0.97 & 25667 \\
				B-LOC & 0.86 & 0.86 & 0.86 & 636 \\
				I-LOC & 0.89 & 0.84 & 0.86 & 5368 \\
				accuracy &  &  & 0.95 & 31671 \\
				macro avg & 0.91 & 0.89 & 0.90 & 31671 \\
				weighted avg & 0.95 & 0.95 & 0.95 & 31671 \\
				\bottomrule
			\end{tabular*}
		\end{minipage}
	\end{center}
\end{table*}

\begin{figure*}
	\centering
		\includegraphics[width=12cm,height=7cm]{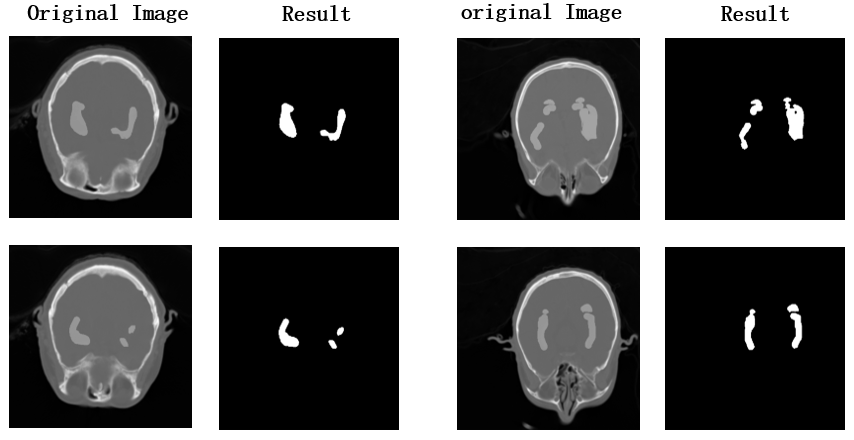}
	\caption{Original Image represents an unprocessed image, Result representing the results of the experiment.}
	\label{Fig.res}
\end{figure*}

\section{Experiment}
In order to further study the dataset, we use our image data using the U-net segmentation model. Due to the unique structure of the U-net model, which aids in capturing the contextual information of images, the left side of U-net is a deep convolutional network responsible for extracting image features. The right side involves an upsampling process that gradually restores the image resolution. Furthermore, U-net excels in handling detailed aspects of images, enabling precise segmentation of small structures within images, which is crucial for medical imaging\cite{zhou2018unet++,oktay2018attention}. We utilized Python scripting to convert NIfTI format images into PNG format. Subsequently, the images were randomly divided into training and testing datasets in a 7:3 ratio to validate the efficacy of image data when utilized with the U-net model. The results of the segmentation are presented in Figure \ref{Fig.res}. We use our text data on a Named Entity Recognition(NER) model, Named Entity Recognition (NER) models represent a key task in natural language processing. The objective of NER is to automatically identify named entities in text and categorize them into predefined classes\cite{skeppstedt2014automatic}. We divided the text data into training, testing and validation sets. After training,the weight files were imported into the test and validation sets to evaluate the effectiveness of the text data, as shown in Table \ref{tab1}. Overall, the experimental validation of the dataset on the U-net segmentation model and NER model proved to be effective. For the image data, it accurately segmented the annotated regions, while the text data successfully extracted the annotated key content. The dataset demonstrated itseffec- tiveness in deployment on deep learning models and achieved favorable results, contributing to the advancement of intelli- gent healthcare to a certain extent\cite{baker2017internet,tian2019smart,catarinucci2015iot,chui2017disease}.

\section{Conclusions}
This paper presents a meticulously annotated dataset tailored for advancing research in the relatively underrepresented domain of HICH image and associated case text information extraction. The dataset is applicable for hot-topic issues like image segmentation and natural language processing, providing a wealth of experimental data for the development of medical image processing and case text analysis algorithms. To demonstrate the versatility of the dataset, we provide an overview of its origin, format, and composition, among other relevant details. We believe that this dataset holds the potential for excellent performance with more refined models and advanced algorithms, contributing significantly to the development of intelligent healthcare solutions. This is expected to enhance the accuracy and efficiency of clinical diagnoses and treatments.

We aimed to update this dataset and expand both the quantity and variety of the dataset regularly. Progress and results were reported in due course.

\bibliographystyle{unsrt}
\bibliography{www}

\end{document}